\documentclass[sigconf,natbib=true,anonymous=false]{acmart}
\usepackage{graphicx}
\usepackage{subfigure} 
\usepackage{multirow}
\usepackage{subcaption}

\usepackage{adjustbox} 
\usepackage{comment}
\usepackage{url}
\usepackage{graphicx}
\usepackage{paralist}
\usepackage{url}
\usepackage{xspace}
\usepackage{amsfonts,amsmath}
\usepackage{booktabs}
\usepackage{setspace}
\usepackage{multirow}
\usepackage{tabularx}
\usepackage{multicol}
\usepackage{wrapfig}
\usepackage{lipsum}
\usepackage{booktabs}
\usepackage{makecell}
\usepackage{array}
\usepackage{amsmath}
\usepackage{xcolor}
\usepackage{latexsym}
\usepackage{pifont}
\usepackage{mdframed}
\usepackage{multirow}
\usepackage{hyperref}
\usepackage{tcolorbox}
\usepackage{booktabs}
\AtBeginDocument{%
  \providecommand\BibTeX{{%
    \normalfont B\kern-0.5em{\scshape i\kern-0.25em b}\kern-0.8em\TeX}}}

\setcopyright{iw3c2w3g}

\acmDOI{https://arxiv.org/abs/2307.11278} 
\acmConference[arxiv]{arxiv}{March 2024}{arxiv} 
\acmYear{2024}
\acmBooktitle{arxiv} 

\begin{document}

\title{AMuRD: Annotated Arabic-English Receipt Dataset for Key Information Extraction and Classification}

\author{Abdelrahman Abdallah}
\email{Abdelrahman.Abdallah@uibk.ac.at}
\affiliation{%
  \institution{University of Innsbruck}
  \streetaddress{Innrain 52}
  \city{Innsbruck}
  \country{Austria}
}
\author{Mahmoud Abdalla}
\email{mahmoud.abdallah@discoapp.ai}
\affiliation{%
  \institution{DISCO AI}
  \streetaddress{Ard El-Marwaha, El-Katamey,}
  \city{Cairo}
  \country{Egypt}
}
\author{Mohamed Elkasaby}
\email{m.abdallah@discoapp.ai}
\affiliation{%
  \institution{DISCO AI}
  \streetaddress{Ard El-Marwaha, El-Katamey,}
  \city{Cairo}
  \country{Egypt}
}
\author{Yasser Elbendary}
\email{yelbendary@discoapp.ai
}
\affiliation{%
  \institution{DISCO AI}
  \streetaddress{Ard El-Marwaha, El-Katamey,}
  \city{Cairo}
  \country{Egypt}
}

\author{Adam Jatowt}
\email{adam.jatowt@uibk.ac.at}
\affiliation{%
  \institution{University of Innsbruck}
  \streetaddress{Innrain 52}
  \city{Innsbruck}
  \country{Austria}
}
\renewcommand{\shortauthors}{Trovato and Tobin, et al.}

\begin{abstract}
The extraction of key information from receipts is a complex task that involves the recognition and extraction of text from scanned receipts. This process is crucial as it enables the retrieval of essential content and organizing it into structured documents for easy access and analysis. In this paper, we present AMuRD, a novel multilingual human-annotated dataset specifically designed for information extraction from receipts. This dataset comprises $47,720$ samples and addresses the key challenges in information extraction and item classification - the two critical aspects of data analysis in the retail industry. Each sample includes annotations for item names and attributes such as price, brand, and more. This detailed annotation facilitates a comprehensive understanding of each item on the receipt. Furthermore, the dataset provides classification into $44$ distinct product categories. This classification feature allows for a more organized and efficient analysis of the items, enhancing the usability of the dataset for various applications. In our study, we evaluated various language model architectures, e.g., by fine-tuning LLaMA models on the AMuRD dataset. Our approach yielded exceptional results, with an F1 score of 97.43\% and accuracy of 94.99\% in information extraction and classification, and an even higher F1 score of 98.51\% and accuracy of 97.06\% observed in specific tasks. The dataset and code are publicly accessible for further research\footnote{\url{https://github.com/Update-For-Integrated-Business-AI/AMuRD}}.
\end{abstract}


\keywords{Receipt extraction, Multilingual,  Arabic, English, Dataset, Information Extraction}


\maketitle

\section{Introduction}
\begin{figure}[h]
    \centering
    \includegraphics[width=0.5\textwidth]{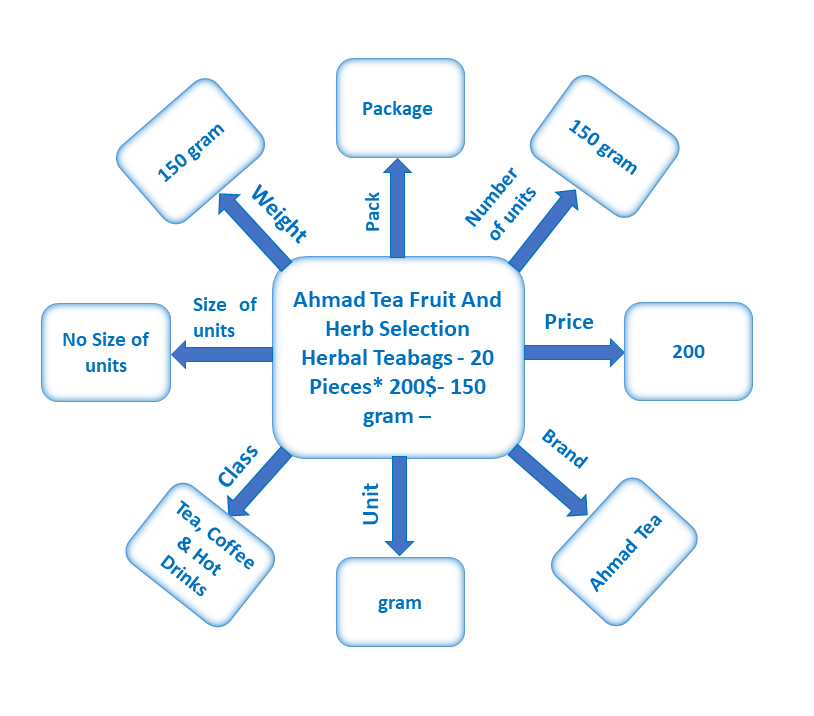}
    \caption{Example of Information Extraction from a Receipt Item}
    \label{fig:exampleindataset}
\end{figure}

Receipt extraction \cite{le2019deep,kasem2023customer,abdallah2023exploring,karatzas2013icdar,karatzas2015icdar,karatzas2018robust} is a task with broad implications for automating business processes, enhancing financial analysis, and enabling efficient inventory management. By capturing and extracting information from scanned receipts, organizations can streamline operations, gain valuable insights, and make informed decisions. 
Consider the example of restaurant receipts. From these, one can extract details such as the dates of the transactions, the list of ordered items (including their names and prices), and more. The ability to accurately extract such information is crucial for businesses in several ways: it enables Expense Tracking, facilitates Financial Analysis, and aids in the Automation of Business Processes. Figure \ref{fig:exampleindataset} shows an example of a key information extraction task that involves recognizing and extracting the information from the item name. The ability to accurately extract such information is crucial for businesses to track expenses, analyze consumer preferences, and manage inventory effectively.

The effectiveness of receipt extraction systems crucially depends on the availability of high-quality datasets that mirror the complexities found in real-world receipts. However, the creation of such datasets is a challenging task due to the diverse nature of receipts, which can vary greatly in terms of layout, language, and information density.

In this paper, we present a novel and extensive dataset, called \textbf{AMuRD} (\textbf{A}nnotated \textbf{Mu}ltilingual \textbf{R}eceipt \textbf{D}ataset), tailored to the task of receipt extraction. This dataset has been constructed with two primary objectives: key information extraction and item classification from scanned receipts. While existing datasets have significantly contributed to the field, the need for multilingual datasets that encompass diverse linguistic and contextual nuances remains pressing. Our dataset addresses this gap by providing a collection of receipts in both Arabic and English, offering a versatile resource for the advancement of information extraction techniques.

AMuRD encompasses $47,720$ samples from a wide array of sources, including retail stores, restaurants, and supermarkets.
To facilitate granular analysis and information extraction, human annotations have been provided for various receipt fields. These annotations include item names, classes, weight, number of units, total price, packaging details, units, and brand names. Our dataset includes $37,419$ unique item names and enables classification into $44$ distinct product categories, enhancing the organization of items. It further provides detailed information on weight, number of units, size of units, price, and total price, offering insights into quantitative aspects of purchased items. By capturing packaging information and presence indicators, researchers gain the ability to study consumer behavior, pricing trends, and promotional strategies within receipts.

Receipt extraction in multilingual settings, such as Arabic and English, introduces unique challenges stemming from such documents' inherent complexity and variability. In the subsequent sections, we explore some of these key challenges and discuss their implications on the accuracy and efficiency of information extraction systems in this domain.

To sum up our contributions in this paper are as follows:
\begin{enumerate}
    \item We introduce \textbf{AMuRD}, a novel multilingual dataset tailored for receipt extraction. This dataset encompasses both Arabic and English content, addressing the pressing need for datasets that cater to diverse linguistic contexts. The AMuRD dataset is extensive, comprising $47,720$ samples from a wide array of sources, ensuring its richness and practical relevance.
    \item Our dataset focuses on two challenges in receipt understanding: key information extraction and item classification. It includes annotations for various receipt fields, enabling understanding of each item on the receipt. Furthermore, it enables classification into $44$ distinct product categories, enhancing the organization of items.
    \item We tested several language model architectures and we have also fine-tuned a specific model, which we refer to as LLama. This fine-tuned LLama model has been tested on AMuRD, achieving an F1 score of $0.76$ and an accuracy of $0.68$. These results demonstrate the effectiveness of our approach in extracting key information and classifying items accurately.
\end{enumerate}


\begin{table*}[t]
\centering
\begin{adjustbox}{width=0.9\textwidth}
\begin{tabular}{c|c|c|c|c|c|c|c|c|c}
\toprule
Dataset & \#Classes & Source & Public & Task & \#Images/Text & Language & KE level & Multilingual & Item-level Annotation\\
\midrule

Intellix & 10 & unknown & Private & IE &12,000 Images & English & Full receipt & No & No\\
CloudScan & 8 & invoice & Private  &IE& 326,471 Images &English  & Full receipt & No & No\\
CUTIE & 9 & receipt & Private & IE& 4,484 Images & Spanish   & Full receipt & No & No\\
SROIE & 4 & receipt & Public/Private &IE& 600 Images & English  & Full receipt & No & No\\
WildReceipt & 25 & receipt & Public &IE & 1,768 Images/50,000 Receipt Text& English  & Full receipt & No & No\\
\textbf{AMuRD} & \textbf{44} &\textbf{receipt} & \textbf{Public} & IE/C & \textbf{47,720 Text} & Arabic/English  &  Items & \textbf{Yes} & \textbf{Yes}\\
\bottomrule
\end{tabular}
\end{adjustbox}

\caption{Comparison between different datasets. "IE" means Information Extraction, and "C" means Classification. The last two columns highlight the unique features of the AMuRD dataset: its multilingual nature and item-level annotation.}

\label{tab:experiment_results}
\end{table*}

\section{Related Work}

In this section, we provide an overview of the existing research and efforts in the field of Information Extraction, with a particular focus on receipt extraction. While Information Extraction from scanned receipts plays a key role in numerous document analysis applications and holds great commercial potential, there has been relatively limited research and advancement in this specific area. 

Previously, several key information extraction datasets were developed, but almost all are private and unavailable to the public. For instance, Intellix \cite{schuster2013intellix} was trained on 8,000 scanned documents with 10 annotated semantic categories. CloudScan \cite{palm2017cloudscan} was tested on 326,471 scanned UBL invoices. CUTIE \cite{zhao2019cutie} annotated 4,484 Spanish receipt documents captured in the wild, including taxi receipts, Meals Entertainment receipts, and hotel receipts, with 9 different key information classes.

WildReceipt \cite{sun2021spatial} is a new public dataset of 1,768 receipt images with 25 classes. It includes both key and value categories, making it a valuable resource for key information extraction research.

The ICDAR 2019 Competition on Scanned Receipt OCR and Information Extraction (SROIE) \cite{huang2019icdar2019} plays important roles for many document analysis research endeavours and applications. SROIE's corresponding ground truth is publicly available. However, the test set ground truth has not been released. This competition aimed to advance Scanned Receipt OCR and Information Extraction and offered three distinct tasks: Scanned Receipt Text Localisation, Scanned Receipt OCR, and Key Information Extraction from Scanned Receipts. 

Even though the SROIE competition was an important step forward, in general, little research has been done in this area, and there is still a lot to improve, especially when it comes to dealing with receipts in different languages. Receipts often have text in multiple languages (e.g., English-Arabic receipts in the Middle East and English-German receipts in Germany \& Austria, and so on), making the task of information extraction challenging. 

In addition to the works mentioned above, other notable efforts in the field of Information Extraction have been taken. For instance, the work by \cite{do2019neural} proposed a novel neural model to identify nested entities by dynamically stacking flat Named Entity Recognition (NER) layers. Each flat NER layer is based on the state-of-the-art flat NER model that captures sequential context representation with a bidirectional Long Short-Term Memory (LSTM) layer and feeds it to the cascaded Conditional Random Field (CRF) layer. This model allows for the extraction of outer entities by taking full advantage of information encoded in their corresponding inner entities, in an inside-to-outside way. The model dynamically stacks the flat NER layers until no outer entities that contain other entities within them are extracted. This approach has been shown to outperform state-of-the-art feature-based systems on nested NER.

Similarly, \citet{liu2020survey} provided a comprehensive survey of the recent advances in Information Extraction, including techniques like deep learning and reinforcement learning. The authors explained the basic concepts of Information Extraction and Deep Learning, primarily expounding on the research progress and achievements of Deep Learning technologies in the field of Information Extraction. They also summarized the prospects and development trends in Deep Learning in the field of Information Extraction as well as difficulties requiring further study. 

There have been other significant works in the field of information extraction. For instance, a survey on Open information Extraction~\cite{niklaus-etal-2018-survey} provides a detailed overview of the various approaches that have been proposed to date to solve the task of Open Information Extraction. Another work by Zheng et al~\cite{zheng-etal-2019-boundary} proposed a boundary-aware neural model for nested NER which leverages entity boundaries to predict entity categorical labels.
 
In our work, we not only build upon the existing efforts in the field of Information Extraction but also introduce novel contributions that set our work apart. Unlike previous works that primarily focus on English text, our dataset is unique in its multilingual nature, encompassing a large number of items in both Arabic and English.  Moreover, the receipts in our dataset have been annotated by experts with key information such as prices and product names. Furthermore, our dataset goes beyond just providing raw data. It aims to enhance the understanding of products, their quantities, and other characteristics listed on the receipts. This focus on understanding the semantics of receipt data is a novel aspect of our work and opens up new possibilities for advanced analytics and insights. In summary, our work stands out in its multilingual coverage, expert annotation, and emphasis on semantic understanding, making it a novel and significant contribution to the field of Information Extraction from receipts

\begin{figure}
    \centering		
    \subfigure[]{
        \includegraphics[width=0.20\textwidth]{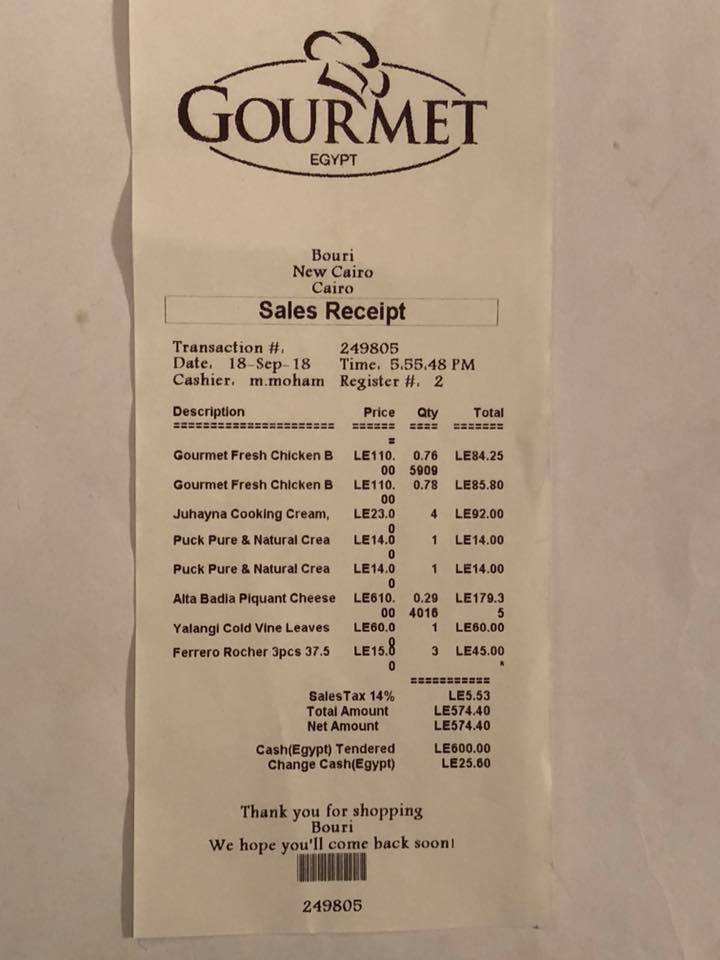}
        \label{fig:net-example1}
    } 
    \hspace{5mm} 
    \subfigure[]{
        \includegraphics[width=0.20\textwidth]{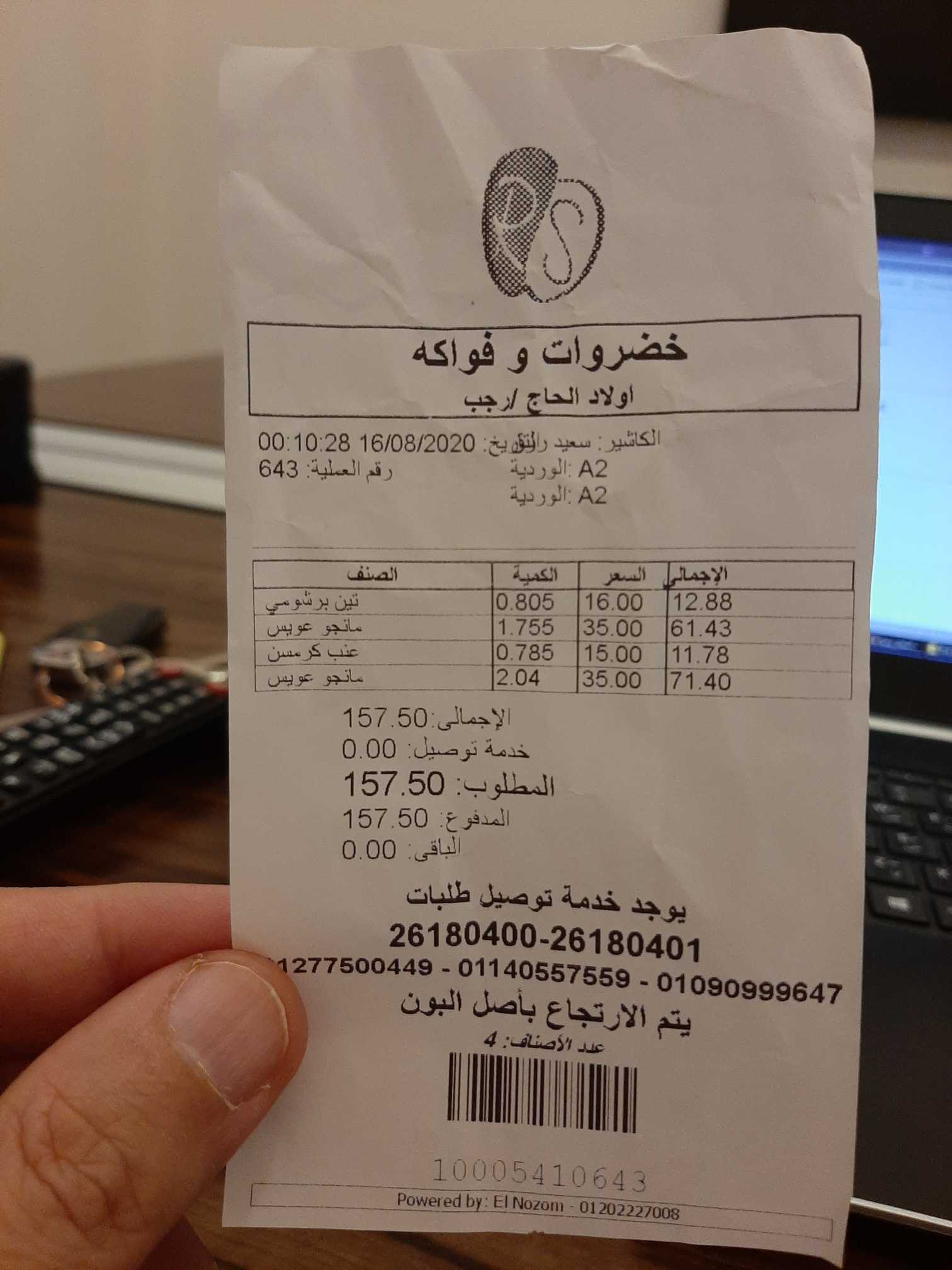}
        \label{fig:net-example2}
    } 
    \caption{Examples of receipts in our dataset.}
    \label{fig:examples_receipts}	 		
\end{figure}

\begin{figure*}[t!]
    \centering
    \includegraphics[width=1.1\textwidth]{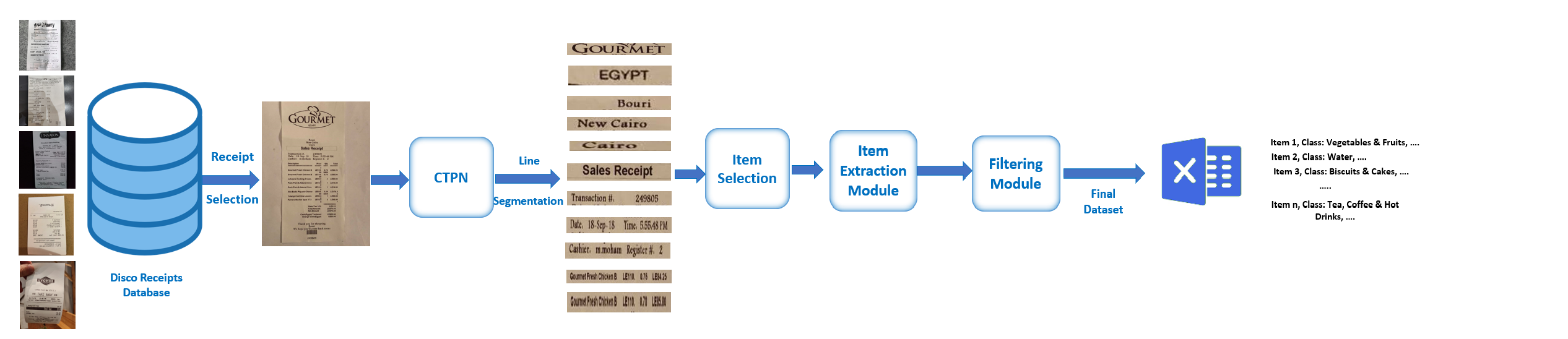}
    \caption{The Workflow of the AMuRD Dataset Framework. The process begins with the receipt Selection Module, where relevant receipts are chosen. These are then passed to the item selection module to select the items from the receipt. The item extraction Module extracts keywords from the items selection, and lastly Filtering Module removes and reviews each annotated item.}
    \label{fig:overflow}
\end{figure*}

\section{Dataset and Annotations}
To create a high-quality dataset for the extraction of key information from receipts in both Arabic and English, we followed a robust and comprehensive methodology. This methodology was designed to ensure the creation of a dataset that is both diverse and representative of real-world receipts. Below we describe the steps we took.

\subsection{Data Collection}
Our first step was to gather a large and diverse collection of receipts. To do this, we utilized the DISCO application\footnote{\url{https://discoapp.ai/}}, a platform that allows registered users to upload real receipts from a variety of commercial establishments, including restaurants, supermarkets, and retail stores. We collected a total of one hundred thousand receipts, carefully chosen to cover a wide range of industries, products, and transaction types. This approach ensured that our dataset was not only rich and relevant but also accurately reflected the complexity and variability of real-world transactional documents. Figure \ref{fig:examples_receipts} provides a visual representation of the diversity of receipts included in our dataset.

\subsection{Annotation Guidelines Development and Annotator Team} To ensure consistency and precision in the annotation process, we developed comprehensive guidelines. These guidelines provide clear instructions and definitions for each element that needs to be annotated, including item names, classes, weight, number of units, size of units, price, total price, packaging, units, and brand names.

Our annotation team consisted of five expert annotators. Each annotator was provided with these guidelines to ensure a uniform process of annotation across the entire dataset. The use of a small, dedicated team of annotators helped to maintain high standards of accuracy and consistency in the annotations. This rigorous approach to annotation has contributed to the high quality of our dataset.

\subsection{Receipt Selection}
Once we had our collection of receipts, the annotators carefully examined each one against several criteria. These criteria included the \emph{clarity of the text}, the \emph{completeness of the receipt image}, and the \emph{character of the document} (it had to be a receipt and not an unrelated bill, such as an electricity bill). This selection ensured that the subsequent annotation process was based on high-quality images and relevant documents.

\subsection{Connectionist Text Proposal Network (CTPN)}
In this step, each receipt was processed through the Connectionist Text Proposal Network (CTPN)~\cite{tian2016detecting}, a tool known for its efficacy in detecting text on a line-by-line basis across various scales and languages, without necessitating further post-processing. Its innovative approach, diverging from traditional bottom-up methods that require extensive post-filtering, reaches an F-measures of 0.88 and 0.61 on the ICDAR 2013 and 2015 benchmarks~\cite{karatzas2013icdar,karatzas2015icdar}, respectively. This step was crucial for preparing each receipt for detailed annotation by isolating every line of text for further analysis.
\subsection{Item Selection}
After a receipt passed the initial screening, it was then passed on to annotators. The annotators went through each line item on the receipt, carefully transcribing every entry. This process prepared the data for the next phase of processing.
\subsection{Item Extraction}
This phase involved expert annotators in the process of information extraction. They took the itemized data and processed it, labeling and classifying each element. This stage is crucial as it transforms the raw, annotated data into structured information that is suitable for further analysis.

\subsection{Data Validation}
The final step in our annotation process was overseen by the head of the annotator team. This individual checked the accuracy of the information extraction and classification. If any discrepancies or errors were identified, the previous steps were reviewed. This rigorous validation process ensured the integrity and reliability of our dataset.

AMuRD consists of $47,720$ items, encompassing both Arabic and English content, and serves as a comprehensive resource for key information extraction. 
Each receipt image contains approximately critical text fields, including item names, unit prices, and total prices. For a visual representation, refer to Figure \ref{fig:examples_receipts}, which showcases sample receipt images.

\subsection{Dataset Statistics}

\subsubsection{Language Distribution}
\begin{figure}[h]
    \centering
    \includegraphics[width=0.3\textwidth]{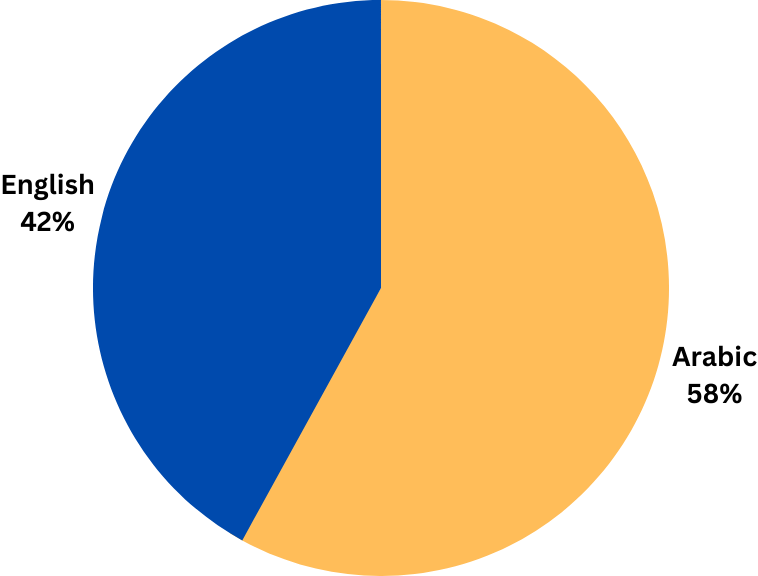}
    \caption{Language Distribution in Item Names}
    \label{fig:language}
\end{figure}

The language distribution in the dataset, as depicted in Figure \ref{fig:language}, reveals the composition of the data. 
This distribution reflects the multilingual nature of real-world receipts, where items and information may be presented in different languages. 

\begin{figure*}[h]
    \centering
    \includegraphics[width=.75\textwidth]{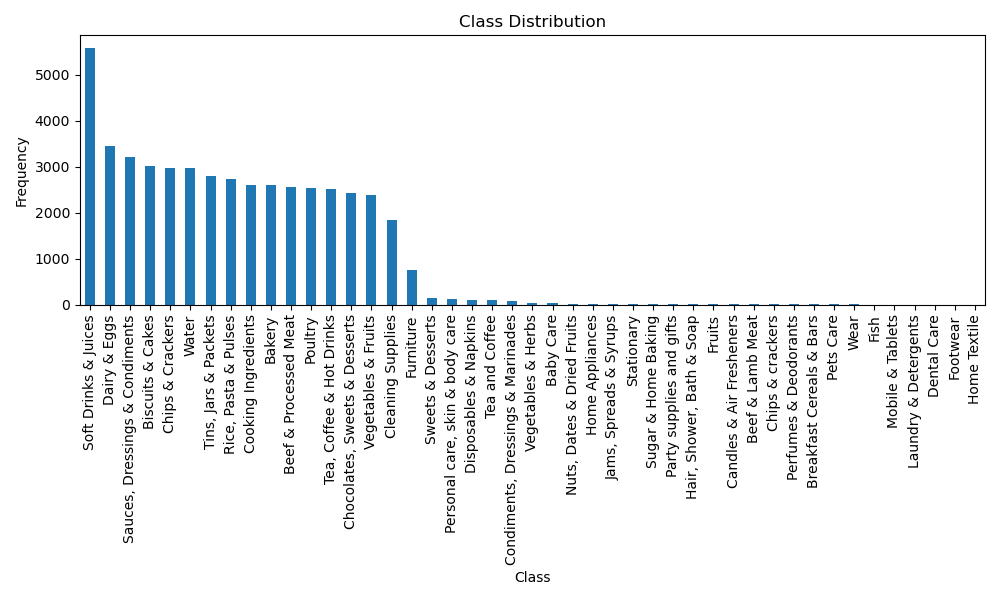}
    \caption{Class Distribution}
    \label{fig:class_distribution}
\end{figure*}

\subsubsection{Class Distribution}
The class distribution in the dataset, as illustrated in Figure \ref{fig:class_distribution}, provides insights into the composition of items across different product categories. 

The dataset encompasses a wide array of product categories, ranging from \textit{Soft Drinks \& Juices} to \textit{Home Textile}. This diversity reflects the richness of products found in real-world receipts, catering to various consumer needs and preferences. Certain categories show a higher frequency of occurrence in the dataset, suggesting their popularity and frequent inclusion in receipts. On the other hand, categories like \textit{Dental Care, Footwear, and Mobile \& Tablets} have relatively low representation, indicating their less frequent occurrence in the dataset. 

It is important to note that some categories have a significantly lower frequency compared to others. This class imbalance can pose challenges for machine learning models, particularly when building classifiers. Techniques such as oversampling, undersampling, or using class weights may be necessary to address this issue.

\subsubsection{Price Distribution}
Analyzing the price distribution within the dataset allows for gaining insights into the economic aspects of the items it contains. The dataset's prices exhibit a wide range, from relatively low-cost items to more expensive ones. 

\begin{figure}[h]
    \centering
    \includegraphics[width=0.4\textwidth]{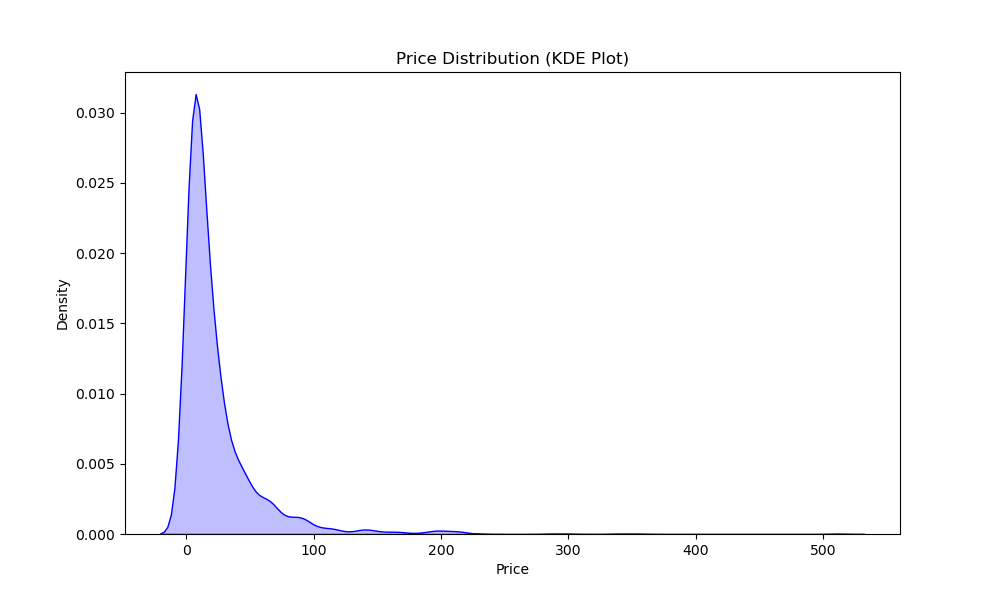}
    \caption{Price Distribution}
    \label{fig:price_distribution}
\end{figure}

The mean price across all items in the dataset is approximately $36.69$, indicating the average cost of items in the dataset. The median price, which is $23.95$, represents the middle point of the price range when all prices are sorted in ascending order. 
The lowest recorded price in the dataset is $0.25$, 
while the dataset includes items with prices as high as $512.00$, indicating the existence of relatively expensive products as well. 
The KDE plot price distribution is given in Fig. \ref{fig:price_distribution}.

\section{Experimental Setup}
In this section, we elaborate on the methodologies employed for item classification and information extraction from the AMuRD dataset. Our approach is multifaceted, involving preprocessing of the raw data, feature extraction, model training, evaluation, and the application of advanced LLM fine-tuning techniques.

\subsection{Item Classification}
Item classification from receipts presents unique challenges due to the wide variety of item names, and categories, and the multilingual nature of the data. Our methodology to tackle these challenges comprises preprocessing, feature extraction, model training, and evaluation.

\subsubsection{Preprocessing}
The preprocessing step is necessary as it prepares the raw data for subsequent analysis. Given the multilingual nature of our dataset, we perform language-specific preprocessing for Arabic and English texts. 

For Arabic text, we apply light stemming using the Farasa stemmer\footnote{\url{https://github.com/MagedSaeed/farasapy}}. Light stemming is a technique used in natural language processing to reduce inflected (or sometimes derived) words to their word stem or root form. The Farasa stemmer is a fast and accurate Arabic light stemmer and tokenizer, making it an excellent choice for our dataset.

For English text, we apply tokenization and lemmatization using the NLTK library \cite{bird2009natural}. 
We also remove stop words in both languages to reduce noise in the data. 

\subsubsection{Feature Extraction}
Feature extraction is a process of dimensionality reduction by which an initial set of raw data is reduced to more manageable groups for processing. In our case, we use the Bag of Words (BoW) model to convert text data into numerical feature vectors. In this model, a text is represented as the bag (multiset) of its words, disregarding grammar and word order but keeping multiplicity. We also experiment with TF-IDF weighting to reflect how important a word is to a document in the corpus. TF-IDF stands for Term Frequency-Inverse Document Frequency, a numerical statistic used to reflect how important a word is to a document in a collection or corpus.

\subsubsection{Models}
Our model training included benchmarks with traditional machine learning algorithms such as Naive Bayes, k-nearest neighbor (KNN), Support Vector Machines (SVM), GradientBoost, and Random Forest to establish simple baselines. In addition, we explored the capabilities of various Large Language Models (LLMs) for item classification, including llama v1, llama v2, Mistral, Mixtral, and GPT-3.5. Importantly, we applied fine-tuning to llama v1 and llama v2 models to enhance their performance on our dataset. We will give more detail about the training in Section \ref{section:training}.

\subsection{Information Extraction}
Information extraction from receipts is critical for transforming unstructured data into actionable insights. We deployed a suite of state-of-the-art models to address the intricacies of this task.

\subsubsection{LLaMA V1 and LLaMA V2}
\label{section:training}

LLaMA V1\cite{touvron2023llama} and LLaMA V2 \cite{touvron2023llamav2} represent two language models from the collection of foundation language models ranging from 7B to 65B parameters. These models are trained on trillions of tokens, all while using publicly available datasets exclusively. LLaMA V1 with $7$ billion to $65$ billion parameters, demonstrates that it is possible to achieve state-of-the-art language modeling without relying on proprietary or inaccessible datasets. LLaMA V1 employs byte pair encoding (BPE) \cite{sennrich2015neural} for tokenization. Key features include pre-normalization with RMSNorm \cite{zhang2019root}, the SwiGLU activation function \cite{shazeer2020glu}, and rotary positional embedding \cite{su2021roformer}. This iteration laids the foundation for future advancements and outperformed larger models like GPT-3.5 $(175B)$ \cite{brown2020language} on various benchmarks.

Both LLaMA V1 and V2 excel in their fine-tuning approaches. LLaMA V1 employs Supervised Fine-Tuning (SFT) \cite{chung2022scaling} with careful annotation and auto-regressive training objectives. LLaMA V2 raises the fine-tuning process with Reinforcement Learning with Human Feedback (RLHF), introducing novel techniques for data collection and reward modeling. It optimizes for both helpfulness and safety, effectively balancing the trade-off between these critical aspects.

\subsubsection{Training \& Prompting}
\label{sec:mod-prompt}

Instruction Tuning \cite{wei2021finetuned,ouyang2022training} is the technique of fine-tuning Large Language Models with multi-task datasets presented in natural language descriptions. We fine-tuned the LLaMA 7.6B \cite{touvron2023llama} model on an instruction-tuning converted AMuRD dataset for keyword information extraction.

The model is trained with supervision using the standard objective of predicting the next token given the previous ones. The dataset includes \texttt{instruction}, \texttt{input}, \texttt{output} fields. For such cases, we construct the prompt:

\begin{quote}
"Below is an instruction that describes a task, paired with an input that provides further context."
"Write a response that appropriately completes the request."
"\#\#\# Instruction:\{instruction\}\#\#\# Input:\{input\}\#\#\# Response:"
\end{quote}

At inference time, the same prompt is used to generate the answer. Only the text generated after \textit{\#\#\# Response:} is used as the final output. We sample from the model using \textit{top-p} sampling \cite{topp} with a temperature of 0.2, $p=0.75$, $k=50$, and beam search with 4 beams. Our experiments will demonstrate the advantage of generic and specific instruction tuning to LLaMA’s keyword extraction and classification ability. There are also existing works that perform multitask instruction tuning of LLMs for specific scenarios, such as machine translation \cite{jiao2023parrot}, information extraction \cite{wang2023instructuie}, and QA \cite{abdallah2023generator}.

\section{Experimental Results}
\subsection{Evaluation Metrics}
We evaluate our models using standard metrics such as precision, recall, F1-score, and accuracy. 
By using these metrics, we ensure a comprehensive and robust evaluation of our models’ performance.
Our dataset was split into three subsets: a test set comprising 4,773 samples, a training set with 38,177 samples, and a validation set containing 4,772 samples. 
\subsection{Experiment Configuration}
This section outlines the experimental setup used for training the LLaMA V1 and V2 model, leveraging the DeepSpeed framework \cite{deepspeed_zero,deepspeed2}. We conduct fine-tuning using the official LLaMA model with the Huggingface Transformers toolkit \cite{wolf2020transformers}. Our primary goal during training is to fine-tune LLaMa to produce the reference response while minimizing the cross-entropy loss. Considering the time and computational resource limitations, we choose a parameter-efficient fine-tuning approach over full-model fine-tuning for most of our experiments by using DeepSpeed. DeepSpeed offers techniques and automated parameter tuning to optimize training efficiency and memory utilization. We have tailored the training process by utilizing DeepSpeed's configuration options to achieve the best results. We used mixed precision training with bfloat16 (bf16) precision, a technique that accelerates training while preserving model accuracy. We chose the AdamW optimizer \cite{loshchilov2017decoupled} as our optimizer and let DeepSpeed automatically determine its hyperparameters. To regulate the learning rate, we employed the WarmupDecayLR scheduler. All experiments are carried
out on 4 Nvidia A100 48GB GPUs.

\subsection{Item Classification}
We have employed a variety of machine learning algorithms and Large Language Models (LLMs), both in their original and fine-tuned forms, to gauge their efficacy in accurately classifying receipt items.

Traditional machine learning models form our baseline. The Gradient Boost algorithm shows moderate precision and recall, leading to an F1 score of 73.19\% and an accuracy of 70.32\%. The K-Nearest Neighbors (KNN) algorithm significantly improves on this, with both precision and recall just above 90\%, resulting in a similar F1 score and accuracy. Naive Bayes further pushes the boundary with slightly improved scores across all metrics, notably a 92.20\% accuracy. Random Forest classifiers show a robust performance with an F1 score of 94.18\% and an accuracy of 94.32\%, while Support Vector Machine (SVM) tops the traditional models with impressive precision and recall, both slightly over 96\%, and the highest F1 score and accuracy, 96.30\% and 96.43\% respectively. These models establish a strong benchmark for item classification as shown in Table. \ref{tab:experiment_classification}.

The non-fine-tuned LLMs, display varied results. DeciLM has a relatively low precision but higher recall, which is reflected in a modest F1 score of 16.17\% and accuracy of 8.79\%. LLaMA V1 and V2 without fine-tuning show limited precision and recall, resulting in F1 scores of 19.13\% and 8.19\% respectively. Similarly, Mistral, LLaMA V2 (13B), and Mixtral display challenges in achieving high scores, with F1 results not exceeding 31.07\%. GPT-3.5 achieves an F1 score of 21.83\% and an accuracy of 12.25\%, indicating that without fine-tuning, even the most powerful LLMs struggle with the specificity required for item classification on our dataset.

\begin{table}[t]
\caption{Item Classification Results }
\begin{adjustbox}{width=0.4\textwidth}
\begin{tabular}{c|c|cccc}
\toprule
Models &parameters & Precision & Recall & F1 & Acc \\
\midrule
GradientBoost& - & 80.25& 70.32& 73.19 & 70.32 \\
KNN & - &   90.09   &  90.10 & 89.92  &  90.10 \\
Naive Bayes& - & 91.14 &92.20 &    91.23  &  92.20 \\
Random Forest& - & 94.27 & 94.32 &     94.18  &  94.32 \\
SVM & -  &  96.30 & 96.43 &  96.30  &  96.43 \\

\hline
DeciLM \textsuperscript{‡}& 6B    &9.08 & 73.81 &  16.17 &8.79   \\
LLaMA V2\textsuperscript{‡} & 7B  & 19.52 & 5.18 &  8.19  &  4.27 \\
LLaMA V1\textsuperscript{‡} & 7B  & 12.17 &  44.72 &  19.13  & 10.58    \\
Mistral \textsuperscript{‡}& 7B    &15.97 &87.97 &  27.03  & 15.62  \\

LLaMA V2\textsuperscript{‡} & 13B & 15.07 & 14.99 & 15.03  & 8.12  \\

Mixtral \textsuperscript{‡}& 8x7B    & 19.72 & 73.22 &   31.07 &  18.39 \\

GPT-3.5\textsuperscript{‡} & 175B & 18.68 & 26.26&   21.83 &  12.25 \\
\hline

LLaMA V2\textsuperscript{†}& 7B & 96.94 & \textbf{100.0} & 98.44 &  96.94 \\
\textbf{LLaMA V1}\textsuperscript{†}& \textbf{7B} & \textbf{97.06} & \textbf{100.0} & \textbf{98.51} & \textbf{97.06}  \\
\bottomrule
\end{tabular}
\end{adjustbox}
\label{tab:experiment_classification}
\end{table}
\footnotetext[1]{\textsuperscript{†} 1-shot results}
\footnotetext[2]{\textsuperscript{‡} Fine-tuned model}

\subsection{Fined-tuned LLaMA for Information Extraction}

\begin{table*}[t]
\centering
\footnotesize 
\begin{adjustbox}{width=1\textwidth}
\begin{tabular}{c|c|cc|cc|cc|cc|cc|cc|cc|cc|cc|cc|cc}
\toprule
\multirow{2}{*}{Models} & \multirow{2}{*}{parameters} & \multicolumn{2}{c}{Brand} & \multicolumn{2}{c}{Weight} & \multicolumn{2}{c}{\# Units} & \multicolumn{2}{c}{S.Units} & \multicolumn{2}{c}{T.Price} & \multicolumn{2}{c}{Price} & \multicolumn{2}{c}{Pack} & \multicolumn{2}{c}{Units} & \multicolumn{2}{c}{Overall}\\
&  & F1 & Acc & F1 & Acc & F1 &  Acc & F1 & Acc  & F1 & Acc & F1 & Acc & F1 & Acc & F1 & Acc & F1 & Acc\\
\midrule

LLaMA V2& 7B&  97.20 &94.55 &  96.13  & 92.56 & 98.83 & 97.69 &  99.81 &99.62 & 98.14 & 96.35 & 98.33 &  96.73 & 94.39 & 89.37 & 96.10 & 92.49  & 97.39 & 94.92 \\
\textbf{LLaMA V1}& 7B&  \textbf{97.28} & \textbf{94.72} & \textbf{96.38}  & \textbf{93.02} & \textbf{98.83} & \textbf{97.69} & \textbf{99.77} & \textbf{99.56} &\textbf{98.34} &  \textbf{96.75} & \textbf{98.14} & \textbf{96.35} & \textbf{94.19} & \textbf{89.02} & \textbf{96.29}  & \textbf{92.85} & \textbf{97.43}  & \textbf{94.99}\\

\bottomrule
\end{tabular}
\end{adjustbox}
\caption{Results for fine-tuned LLaMA V1 and LLaMA V2 Models}
\label{tab:experiment_results}
\end{table*}
Table~\ref{tab:experiment_results} presents the nuanced capabilities of our fine-tuned models, LLaMA V1 and LLaMA V2, both wielding 7 billion parameters. The comprehensive results across diverse information extraction and classification categories speak to their effectiveness. In discerning brand details, the models exhibit high prowess, with F1 scores soaring above 97\%, reflecting their acute brand recognition capabilities. Weight and quantity details, often challenging due to their variability, are accurately captured by the models, as evidenced by F1 scores near 96\% for weight and nearly 99\% for both number of units and size of units. Such high scores are indicative of the models' precise identification and classification processes.

The adeptness of these models extends to the crucial financial metrics of total price and price, where they perform well, achieving F1 scores and accuracies above 96\%. This demonstrates their capacity to handle the intricacies of numerical data extraction from receipts, which can vary greatly in format. In other intricate categories like packaging and units, the models again prove their merit with F1 scores in the mid-90s and accuracies that reflect a high level of consistency and reliability. Overall, the models achieve remarkable F1 scores of above 97\% and accuracies nearing 95\%, showcasing their comprehensive effectiveness in both information extraction and classification tasks.


\begin{table}[ht]
\caption{Performance Metrics for zero-shot Information Extraction Across Multiple Categories}
\centering
\footnotesize
\begin{adjustbox}{width=0.5\textwidth}
\begin{tabular}{c|c|c|c|c|c|c|c|c}
\toprule
\multirow{2}{*}{Category} & \multirow{1}{*}{Metric} & \multicolumn{1}{c|}{LLaMA V1} & \multicolumn{1}{c|}{LLaMA V2} & \multicolumn{1}{c|}{LLaMA V2} & \multicolumn{1}{c|}{Mistral} & \multicolumn{1}{c|}{Mixtral} & \multicolumn{1}{c|}{Falcon} & \multicolumn{1}{c}{Zephyr}\\
\cline{2-9}
& Parameter & 7B & 7B  & 13B  & 7B & 8x7B & 7B & 7B\\ \hline

\multirow{4}{*}{Brand} & Precision & 3.14 & 21.10 & 18.62 & 21.37 & 20.32 & \textbf{27.94} & 30.34 \\
& Recall & 30.91 & 59.61 & 14.74 & 83.90 & 65.25 & 37.11 & \textbf{86.94} \\
& F1 & 5.70 & 31.16 & 16.46 & 34.07 & 30.98 & 31.88 & \textbf{44.98} \\
& Accuracy & 2.93 & 18.46 & 8.97 & 20.53 & 18.33 & 18.96 & \textbf{29.02} \\

\hline
\multirow{4}{*}{Weight} & Precision & 3.64 & 20.42 & 15.18 & 16.66 & 14.37 & 16.91 & \textbf{22.88} \\
& Recall & 23.31 & 58.87 & 12.48 & 81.49 & 56.70 & 26.35 & \textbf{83.19} \\
& F1 & 6.29 & 30.32 & 13.70 & 27.66 & 22.93 & 20.60 & \textbf{35.89} \\
& Accuracy & 3.25 & 17.87 & 7.35 & 16.05 & 12.95 & 11.48 & \textbf{21.87} \\

\hline
\multirow{4}{*}{\# Units} & Precision & 6.07 & 71.47 & \textbf{80.42} & 75.93 & 61.60 & 50.19 & 69.70 \\
& Recall & 30.72 & 83.31 & 42.95 & 95.60 & 84.76 & 51.42 & \textbf{93.55} \\
& F1 & 10.14 & 76.94 & 56.00 & 84.64 & 71.35 & 50.80 & \textbf{79.88} \\
& Accuracy & 5.34 & 62.52 & 38.89 & 73.37 & 55.46 & 34.05 & \textbf{66.50} \\

\hline
\multirow{4}{*}{S.Units} & Precision & 0.02 & 0.02 & 0.00 & 0.11 & 0.05 & \textbf{0.19} & 0.09 \\
& Recall & 0.14 & 0.17 & 0.00 & 2.60 & 0.41 & 0.39 & \textbf{1.75} \\
& F1 & 0.04 & 0.04 & 0.00 & 0.21 & 0.08 & \textbf{0.25} & 0.17 \\
& Accuracy & 0.02 & 0.02 & 0.00 & 0.10 & 0.04 & \textbf{0.13} & 0.08 \\

\hline
\multirow{4}{*}{T.Price} & Precision & 0.05 & 0.43 & \textbf{0.84} & 0.79 & 0.62 & 0.65 & 0.51 \\
& Recall & 0.24 & 2.87 & 0.75 & 12.90 & 4.55 & 1.32 & \textbf{8.16} \\
& F1 & 0.08 & 0.75 & \textbf{0.79} & 1.50 & 1.08 & 0.88 & 0.96 \\
& Accuracy & 0.04 & 0.38 & \textbf{0.40} & 0.75 & 0.54 & 0.44 & 0.48 \\

\hline
\multirow{4}{*}{Price} & Precision & 0.05 & 0.86 & 0.69 & 1.03 & 1.17 & 0.71 & \textbf{0.82} \\
& Recall & 0.26 & 5.61 & 0.65 & 18.80 & 8.90 & 1.46 & \textbf{13.31} \\
& F1 & 0.08 & 1.50 & 0.67 & 1.95 & 2.07 & 0.96 & \textbf{1.54} \\
& Accuracy & 0.04 & 0.75 & 0.34 & 0.98 & 1.05 & 0.48 & \textbf{0.78} \\

\hline
\multirow{4}{*}{Pack} & Precision & 0.05 & 0.19 & 0.17 & 0.36 & 0.05 & \textbf{0.69} & 0.47 \\
& Recall & 0.24 & 1.29 & 0.16 & 4.35 & 0.36 & 1.21 & \textbf{6.82} \\
& F1 & 0.08 & 0.33 & 0.17 & 0.67 & 0.08 & \textbf{0.88} & 0.88 \\
& Accuracy & 0.04 & 0.17 & 0.08 & 0.34 & 0.04 & \textbf{0.44} & 0.44 \\

\hline
\multirow{4}{*}{Units} & Precision & 0.00 & 2.60 & 6.94 & 10.37 & 2.27 & 0.00 & \textbf{16.51} \\
& Recall & 0.00 & 13.97 & 13.60 & 86.19 & 3.17 & 0.00 & \textbf{71.71} \\
& F1 & 0.00 & 4.39 & 9.19 & 18.52 & 2.65 & 0.00 & \textbf{26.85} \\
& Accuracy & 0.00 & 2.24 & 4.82 & 10.20 & 1.34 & 0.00 & \textbf{15.50} \\

\hline
\multirow{4}{*}{Overall} & Precision & 1.93 & 14.68 & 14.96 & 15.95 & 13.10 & 12.26 & \textbf{17.74} \\
& Recall & 5.68 & 50.06 & 13.40 & 78.64 & 43.89 & 20.12 & \textbf{76.81} \\
& F1 & 2.88 & 22.70 & 14.14 & 26.53 & 20.18 & 15.24 & \textbf{28.82} \\
& Accuracy & 1.46 & 12.80 & 7.61 & 15.29 & 11.22 & 8.25 & \textbf{16.83} \\
\hline
\end{tabular}
\end{adjustbox}
\label{tab:0-shot_result}
\end{table}

\begin{table}[ht]
\caption{Performance Metrics for 1-shot Information Extraction Across Multiple Categories}
\centering
\footnotesize
\begin{adjustbox}{width=0.5\textwidth}
\begin{tabular}{c|c|c|c|c|c|c|c|c}
\toprule
\multirow{2}{*}{Category} & \multirow{1}{*}{Metric} & \multicolumn{1}{c|}{LLaMA V1} & \multicolumn{1}{c|}{LLaMA V2} & \multicolumn{1}{c|}{LLaMA V2} & \multicolumn{1}{c|}{Mistral} & \multicolumn{1}{c|}{Mixtral} & \multicolumn{1}{c|}{Falcon} & \multicolumn{1}{c}{Zephyr}\\
\cline{2-9}
& Parameter & 7B & 7B  & 13B  & 7B & 8x7B & 7B & 7B\\ \hline

\multirow{4}{*}{Brand} & Precision & 38.06 & 38.37 & 33.27 & 26.97 & 32.99 & \textbf{44.41} & 38.81 \\
& Recall & 69.83 & 43.79 & 62.83 & 88.34 & \textbf{94.14} & 36.47 & 54.00 \\
& F1 & 49.27 & 40.9 & 43.51 & 41.32 & 48.86 & 40.05 & \textbf{45.16} \\
& Accuracy & 32.68 & 25.71 & 27.80 & 26.04 & 32.33 & 25.04 & \textbf{29.16} \\
\hline
\multirow{4}{*}{Weight} & Precision & 49.71 & 40.07 & 44.03 & 41.72 & 43.86 & 43.96 & \textbf{47.70} \\
& Recall & 74.28 & 47.23 & 69.33 & 94.04 & \textbf{95.30} & 36.55 & 58.95 \\
& F1 & 59.56 & 43.35 & 53.86 & 57.80 & 60.07 & 39.91 & \textbf{52.73} \\
& Accuracy & 42.41 & 27.68 & 36.85 & 40.65 & 42.93 & 24.93 & \textbf{35.81} \\
\hline
\multirow{4}{*}{\# Units} & Precision & 33.45 & 67.95 & \textbf{75.76} & 65.65 & 22.86 & 32.80 & 36.39 \\
& Recall & 64.36 & 60.17 & \textbf{79.50} & 94.40 & 91.04 & 30.12 & 52.29 \\
& F1 & 44.02 & 63.82 & \textbf{77.58} & 77.44 & 36.54 & 31.40 & 42.92 \\
& Accuracy & 28.22 & 46.87 & \textbf{63.38} & 63.19 & 22.35 & 18.63 & 27.32 \\
\hline
\multirow{4}{*}{S.Units} & Precision & 77.44 & 43.07 & 62.16 & 56.80 & 67.30 & \textbf{98.75} & 85.25 \\
& Recall & 79.22 & 48.18 & 76.09 & 93.62 & 96.70 & 56.47 & \textbf{71.90} \\
& F1 & 78.32 & 45.48 & 68.42 & 70.70 & 79.36 & \textbf{71.85} & 78.01 \\
& Accuracy & 64.36 & 29.44 & 52.00 & 54.68 & 65.79 & \textbf{56.07} & 63.94 \\
\hline
\multirow{4}{*}{T.Price} & Precision & 93.17 & 76.73 & 84.29 & 83.94 & 91.28 & \textbf{94.50} & 94.05 \\
& Recall & 81.46 & 63.13 & 79.95 & 95.91 & 97.60 & 55.38 & \textbf{73.79} \\
& F1 & 86.92 & 69.27 & 82.06 & 89.53 & 94.33 & 69.84 & \textbf{82.70} \\
& Accuracy & 76.87 & 52.99 & 69.58 & 81.04 & 89.27 & 53.66 & \textbf{70.50} \\
\hline
\multirow{4}{*}{Price} & Precision & 92.93 & 72.86 & 82.09 & 73.63 & 91.39 & \textbf{94.50} & 93.83 \\
& Recall & 81.64 & 62.31 & 80.61 & 95.67 & 97.60 & 55.38 & \textbf{73.77} \\
& F1 & 86.92 & 67.18 & 81.34 & 83.22 & 94.39 & 69.84 & \textbf{82.60} \\
& Accuracy & 76.87 & 50.58 & 68.55 & 71.25 & 89.38 & 53.66 & \textbf{70.35} \\
\hline
\multirow{4}{*}{Pack} & Precision & 5.40 & 3.45 & 3.16 & 4.13 & 3.89 & 4.73 & \textbf{11.26} \\
& Recall & 20.08 & 7.00 & 10.24 & 49.35 & 60.94 & 5.83 & \textbf{23.12} \\
& F1 & 8.51 & 4.63 & 4.83 & 7.62 & 7.31 & 5.22 & \textbf{15.14} \\
& Accuracy & 4.44 & 2.37 & 2.47 & 3.96 & 3.79 & 2.68 & \textbf{8.19} \\
\hline
\multirow{4}{*}{Units} & Precision & 13.33 & 9.56 & 25.74 & 24.52 & 10.32 & 0.00 & \textbf{37.32} \\
& Recall & 0.04 & 38.05 & 74.52 & 87.26 & 1.60 & 0.00 & \textbf{39.89} \\
& F1 & 0.08 & 15.29 & 38.26 & 38.29 & 2.77 & 0.00 & \textbf{38.56} \\
& Accuracy & 0.04 & 8.28 & 23.65 & 23.67 & 1.40 & 0.00 & \textbf{23.88} \\
\hline
\multirow{4}{*}{Overall} & Precision & 55.56 & 43.02 & 51.32 & 47.19 & 49.73 & 51.73 & \textbf{56.07} \\
& Recall & 60.42 & 51.14 & 72.73 & 92.94 & 77.33 & 40.39 & \textbf{60.71} \\
& F1 & 57.89 & 46.73 & 60.18 & 62.60 & 60.54 & 45.36 & \textbf{58.30} \\
& Accuracy & 40.74 & 30.49 & 43.04 & 45.56 & 43.41 & 29.33 & \textbf{41.15} \\
\hline
\end{tabular}
\end{adjustbox}
\label{tab:1-shot_result}
\end{table}

\begin{table*}[ht]
\caption{Performance Metrics for Few-shot Information Extraction Across Multiple Categories}

\centering
\footnotesize 
\begin{adjustbox}{width=1\textwidth}
\begin{tabular}{c|c|c|cc|cc|cc|cc|cc|cc|cc|cc|cc|cc|cc|}
\toprule
\multirow{2}{*}{Models} & \multirow{2}{*}{parameters} & \multirow{2}{*}{\#Shots} & \multicolumn{2}{c|}{Brand} & \multicolumn{2}{c|}{Weight} & \multicolumn{2}{c|}{\# Units} & \multicolumn{2}{c|}{S.Units} & \multicolumn{2}{c|}{T.Price} & \multicolumn{2}{c|}{Price} & \multicolumn{2}{c|}{Pack} & \multicolumn{2}{c|}{Units} & \multicolumn{2}{c|}{Overall}\\
& & & F1  & Acc  & F1  & Acc & F1 & Acc & F1  & Acc & F1 & Acc & F1  & Acc & F1  & Acc & F1  & Acc & F1  & Acc\\
\midrule
\textbf{LLaMA V1}& 7B & 0 & 5.70 & 2.93 & 6.29 & 3.25 & 10.14 & 5.34 & 0.04 & 0.02 & 0.08 & 0.04 & 0.08 & 0.04 & 0.08 & 0.04 & 0.00 & 0.00 & 2.88 & 1.46 \\
\textbf{LLaMA V2}& 7B & 0 & 31.16 & 18.46 & 30.32 & 17.87 & 76.94 & 62.52 & 0.04 & 0.02 & 0.75 & 0.38 & 1.50 & 0.75 & 0.33 & 0.17 & 4.39 & 2.24 & 22.70 & 12.80 \\
\textbf{LLaMA V2}& 13B & 0 & 16.46 & 8.97 & 13.70 & 7.35 & 56.00 & 38.89 & 0.00 & 0.00 & 0.79 & 0.40 & 0.67 & 0.34 & 0.17 & 0.08 & 9.19 & 4.82 & 14.14 & 7.61 \\
\textbf{Mistral}& 7B & 0 & 34.07 & 20.53 & 27.66 & 16.05 & \textbf{84.64} & \textbf{73.37} & \textbf{0.21} & 0.10 & \textbf{1.50} & \textbf{0.75} & 1.95 & 0.98 & 0.67 & 0.34 & 18.52 & 10.20 & 26.53 & 15.29 \\
\textbf{Mixtral}& 8x7B & 0 & 30.98 & 18.33 & 22.93 & 12.95 & 71.35 & 55.46 & 0.08 & 0.04 & 1.08 & 0.54 & \textbf{2.07} & \textbf{1.05} & 0.08 & 0.04 & 2.65 & 1.34 & 20.18 & 11.22 \\
\textbf{Falcon}& 7B & 0 & 31.88 & 18.96 & 20.60 & 11.48 & 50.80 & 34.05 & 0.25 & \textbf{0.13} & 0.88 & 0.44 & 0.96 & 0.48 & 0.88 & 0.44 & 0.00 & 0.00 & 15.24 & 8.25 \\
\textbf{Zephyr}& 7B & 0 & \textbf{44.98} & \textbf{29.02} & \textbf{35.89} & \textbf{21.87} & 79.88 & 66.50 & 0.17 & 0.08 & 0.96 & 0.48 & 1.54 & 0.78 & \textbf{0.88} & \textbf{0.44} & \textbf{26.85} & \textbf{15.50} & \textbf{28.82} & \textbf{16.83} \\

\midrule
\textbf{LLaMA V1}& 7B & 1 & \textbf{49.27} & \textbf{32.68} & 59.56 & 42.41 & 44.02 & 28.22 & 78.32 & 64.36 & 86.92 & 76.87 & 86.92 & 76.87 & 8.51 & 4.44 & 0.08 & 0.04 & 57.89 & 40.74 \\

\textbf{LLaMA V2}& 7B & 1 & 40.9 & 25.71 & 43.35 & 27.68 & 63.82 & 46.87 & 45.48 & 29.44 & 69.27 & 52.99 & 67.18 & 50.58 & 4.63 & 2.37 & 15.29 & 8.28 & 46.73 & 30.49 \\

\textbf{LLaMA V2}& 13B & 1 & 43.51 & 27.80 & 53.86 & 36.85 & \textbf{77.58} & \textbf{63.38} & 68.42 & 52.00 & 82.06 & 69.58 & 81.34 & 68.55 & 4.83 & 2.47 & 38.26 & 23.65 & 60.18 & 43.04 \\

\textbf{Mistral}& 7B & 1 & 41.32 & 26.04 & 57.80 & 40.65 & 77.44 & 63.19 & 70.70 & 54.68 & 89.53 & 81.04 & 83.22 & 71.25 & 7.62& 3.96 & 38.29 & 23.67 & \textbf{62.60} & \textbf{45.56} \\

\textbf{Mixtral}& 8x7B & 1 & 48.86 & 32.33 & \textbf{60.07} & \textbf{42.93} & 36.54 & 22.35 & \textbf{79.36} & \textbf{65.79} & \textbf{94.33} & \textbf{89.27} & \textbf{94.39} & \textbf{89.38} & 7.31 & 3.79 & 2.77 & 1.40 & 60.54 & 43.41 \\

\textbf{Falcon}& 7B & 1 & 40.05 & 25.04 & 39.91 & 24.93 & 31.40 & 18.63 & 71.85 & 56.07 & 69.84 & 53.66 & 69.84 & 53.66 & 5.22 & 2.68 & 0.00 & 0.00 & 45.36 & 29.33 \\

\textbf{Zephyr}& 7B & 1 & 45.16 & 29.16 & 52.73 & 35.81 & 42.92 & 27.32 & 78.01 & 63.94 & 82.70 & 70.50 & 82.60 & 70.35 & \textbf{15.14} & \textbf{8.19} & \textbf{38.56} & \textbf{23.88} & 58.30 & 41.15 \\

\midrule
\textbf{LLaMA V1}& 7B & 2 & 44.75 & 28.82 & 73.73 & 58.39 & 81.40 & 68.63 & 84.29 & 72.84 & 89.50 & 80.99 & 89.42 & 80.87 & 43.58 & 27.86 & 0.0 & 0.0 & 68.68 & 52.30\\

\textbf{LLaMA V2}&  7B & 2 &  54.17 & 37.14 & 74.19 & 58.97  & \textbf{95.42} & \textbf{91.24} & 82.64 & 70.41 & 96.05 & 92.41 & 96.16  &  92.60 &  55.40 &  38.31 &  52.95 & 36.01 & 78.52  & 64.64\\

\textbf{LLaMA V2}& 13B & 2 & 53.22 & 36.26 & \textbf{76.33} & \textbf{61.72} & 95.06 & 90.59  & \textbf{96.01} & \textbf{92.33} & 97.16 & 94.48 & 97.07  & 94.32 &  15.53 & 8.42 &   47.55 & 31.19 & 77.80  & 63.66\\

\textbf{Mistral}& 7B & 2 & 51.33 & 34.52 & 75.29 & 60.38 & 93.48 & 87.76 & 91.79 & 84.83  &  96.58 & 93.40 &  96.74 & 93.69 &  43.66  & 27.92 &  49.71 & 33.08 & 78.38  & 64.45\\

\textbf{Mixtral}& 8x7B & 2 &45.35  & 29.33 & 65.07 & 48.22 &  80.68 & 67.63 &  71.60 & 55.77 & 86.00 & 75.44 &  85.86 & 75.23 & 12.12  & 6.45 & 29.41  & 17.24 & 63.86  & 46.91 \\

\textbf{Falcon}& 7B & 2 & 49.95 & 33.29 & 67.01 & 50.38 & 83.85 & 72.19 & 91.78 & 84.81  & 87.98 & 78.54 & 86.93  & 76.89 & 15.78  & 8.56 & 0.12  & 0.06 & 67.19  & 50.59 \\

\textbf{Zephyr}& 7B & 2 & \textbf{54.43} & \textbf{37.39} & 73.31 & 57.86 & 94.57 & 89.71 & 92.87 & 86.69 & \textbf{96.95} & \textbf{94.09} &  \textbf{96.90} & \textbf{93.98} &  \textbf{59.36} & \textbf{42.21} & \textbf{65.35}  & \textbf{48.54} &  \textbf{81.52} & \textbf{68.81} \\

\midrule
\textbf{LLaMA V1}& 7B & 3 & 46.45 & 30.25 & 66.32 & 49.61 & 85.55 & 74.75 & 92.72 & 86.44 & 91.96 & 85.12 & 91.76 & 84.78 & 39.07 & 24.28 & 0 & 0 & 70.47 & 54.40\\

\textbf{LLaMA V2}& 7B & 3 & 54.43 & 37.39 & 73.31 & 57.86 & 94.57 & 89.71 & 92.87 & 86.69 & 96.95 & 94.09 & 96.90 & 93.98 & \textbf{59.36} & \textbf{42.21} & 65.35 & 48.54 & 81.52 & 68.81\\

\textbf{LLaMA V2}& 13B & 3 & 51.49 & 34.67 & 68.53 & 52.12 & \textbf{95.71} & \textbf{91.78} & \textbf{98.19} & \textbf{96.45} & \textbf{97.17} & \textbf{94.51} & \textbf{97.12} & \textbf{94.40} & 43.07 & 27.44 & 55.69 & 38.59 & 79.69 & 66.25\\

\textbf{Mistral}& 7B & 3 & 53.88 & 36.87 & 78.11 & 64.08 & 94.05 & 88.77 & 96.59 & 93.42 & 97.08 & 94.34 & 96.88 & 93.96 & 51.90 & 35.05 & \textbf{68.85} & \textbf{52.50} & \textbf{82.26} & \textbf{69.87}\\

\textbf{Mixtral}& 8x7B & 3 & 53.65 & 36.66 & 71.71 & 55.89 & 89.24 & 80.57 & 82.13 & 69.68 & 95.28 & 90.99 & 95.15 & 90.76 & 22.40 & 12.61 & 31.61 & 18.77 & 72.60 & 56.99\\

\textbf{Falcon}& 7B & 3 & \textbf{64.30} & \textbf{47.39} & 65.68 & 48.90 & 93.73 & 88.20 & 97.79 & 95.68 & 95.34 & 91.09 & 95.31 & 91.05 & 16.88 & 9.21 & 0.0 & 0.0 & 74.16 & 58.94\\

\textbf{Zephyr}& 7B & 3 & 58.36 & 41.21 & \textbf{74.24} & \textbf{59.04} & 86.20 & 75.75 & 89.64 & 81.22 & 95.93 & 92.18 & 95.86 & 92.05 & 59.22 & 42.06 & 53.67 & 36.68 & 78.80 & 65.02\\

\bottomrule
\end{tabular}
\end{adjustbox}
\label{tab:few_results}
\end{table*}

\subsection{Zero-shot Information Extraction Performance}

The exploration of zero-shot capabilities within our study presents a novel approach to information extraction without direct examples for training. Table~\ref{tab:0-shot_result} showcases the performance metrics of various language models, including LLaMA V1 and V2, under zero-shot conditions across a spectrum of categories. Zero-shot, by design, challenges models to apply learned knowledge from unrelated tasks to make accurate predictions on unseen data, a task that demands a high degree of generalization.

In the context of zero-shot information extraction, the models exhibited varied levels of proficiency across different categories. Notably, in the Brand and Weight categories, the performance metrics indicate a struggle to maintain high precision and recall, which translates to modest F1 scores and accuracy. This trend highlights the inherent challenge of zero-shot performance, where models must rely solely on their pre-existing knowledge base without the benefit of specific training examples.

Among the models, Zephyr and Falcon showed relatively higher F1 scores in several categories, suggesting a better generalization capability in certain contexts. However, LLaMA V1 and V2's performance, while not leading in the zero-shot scenario, still demonstrates the potential of large language models to adapt their extensive pre-trained knowledge to specific tasks.

The Number of Units and Size of Units categories further illustrate the models' challenges in accurately interpreting and extracting numerical data without explicit examples. However, some models like LLaMA V2 (13B) show promising results, indicating a capacity for significant improvement with fine-tuning and targeted training.

The Total Price and Price extraction tasks reveal an interesting aspect of zero-shot learning, where even without specific training, models like LLaMA V1 and V2 can leverage their linguistic and numerical understanding to achieve modest success in extracting financial information.

Packaging (Pack) and Unit categories underscore the complexity of zero-shot information extraction, with generally lower performance across the models because such information is typically not explicitly mentioned in items. Hence, these categories likely suffer from the lack of direct examples that would help models understand the specific context and terminology associated with packaging information and unit measurements.

\subsection{One-shot Information Extraction Results }
Table~\ref{tab:1-shot_result} delineates the performance of various language models in a one-shot learning scenario across multiple categories. In one-shot learning, the models are expected to generalize from a single example to new instances, which is a challenging task, especially in the context of information extraction from diverse datasets like receipts.

When comparing the models' one-shot capabilities, it becomes evident that certain models excel in specific categories. For instance, in the Brand category, Falcon demonstrates the highest precision, while Mixtral leads with the highest recall. However, it is LLaMA V1 that balances both metrics most effectively, achieving the highest F1 score and accuracy overall.

In the Weight category, once again, LLaMA V1 stands out with the highest F1 score and accuracy, suggesting that it has a better understanding of contextual cues related to product weight, despite the varied ways in which this information can be presented in receipts.

The Number of Units and Size of Units categories are particularly demanding since they require the model to interpret potentially ambiguous numerical data accurately. In both categories, LLaMA V2 (13B) shows an impressive recall, but it is LLaMA V1 that maintains a superior balance as reflected in its F1 scores and accuracy.

The Total Price and Price categories are dominated by LLaMA V1 in terms of F1 and accuracy, reinforcing its superior performance in the one-shot extraction of numerical data.

Packaging (Pack) and Units represent categories where the models generally perform less effectively, with Zephyr showing an unexpectedly higher F1 score in Pack, and LLaMA V2 (13B) leading in the Units category.

\subsection{Few-shot Information Extraction Results }

Finally, in few-shot learning, as Table~\ref{tab:few_results} shows, the models are provided with a few examples (2 or 3) from which to learn, allowing them to better understand the nuances of the data.

When examining the performance improvements between one-shot and few-shot scenarios, it is clear that additional examples aid the models significantly. For example, LLaMA V2 (7B) jumps from an F1 score of 46.73\% to 78.52\% in the Overall category when moving from one to two shots, and LLaMA V1 shows similar improvements.

Across specific categories, the advantages of few-shot learning become even more pronounced. In Brand extraction, LLaMA V2 (7B) improves markedly in precision, recall, F1, and accuracy with each additional example provided. In the Weight and Number of Units categories, we see substantial enhancements in F1 scores and accuracies for almost all models, with LLaMA V2 (13B) showing significant jumps in performance.

The Total Price and Price categories also benefit from the few-shot approach, with most models showing noticeable improvements in precision and recall, leading to higher F1 scores. This suggests that with even a small increase in data, the models can better capture the patterns necessary for extracting financial information.

Interestingly, the Pack and Units categories, which were challenging in the one-shot setting, see varied improvements with few-shot learning. While some models like LLaMA V2 (7B) and Mistral exhibit substantial gains, others like Falcon and Zephyr show little to no improvement, indicating that these categories might be inherently more challenging or require a different approach for learning.

\section{Limitations}

This study, while comprehensive in its approach to evaluating the effectiveness of Large Language Models (LLMs) for information extraction from receipts, acknowledges several limitations that could impact the breadth and depth of the findings. One significant limitation is the constrained model input size, which may not fully capture the complexity or entirety of longer receipts. This restriction could potentially limit the models' ability to understand and process the full context of the data, thereby affecting the accuracy of the information extraction.

Furthermore, the scope of models covered in this research is selective, focusing on a specific set of LLMs. Consequently, this study does not represent the full spectrum of models available or those emerging within the rapidly advancing field of natural language processing and machine learning. There are numerous other models, each with unique architectures and capabilities, that were not examined by us. The inclusion of a broader array of models might have yielded different insights or highlighted alternative strengths and weaknesses in the context of receipt information extraction.

In addition to the above limitations related to model selection and input size, this study also acknowledges a limitation concerning the dataset used. The dataset, while extensive and multilingual, does not always represent full receipts. This omission can introduce challenges in understanding the complete transaction context, potentially leading to gaps in the extracted information. The absence of full receipts might limit the models' ability to accurately classify and extract information, as certain relevant details that could influence understanding or classification may be missing.

These limitations highlight areas for future research, including the exploration of models capable of handling larger input sizes and the inclusion of a wider range of LLMs to fully assess the capabilities and limitations of current technology. Additionally, enhancing the dataset to include full receipts or more comprehensive transaction details could significantly improve the quality of information extraction and provide more insights into the models' performance across a broader spectrum of real-world scenarios.

\section{Conclusions}

This work introduced the AMuRD dataset, an extensive multilingual collection tailored for the complex challenges of extracting key information and classifying items from scanned receipts. Comprising 47,720 annotated samples in both Arabic and English, AMuRD paves the way for advanced research and development in the fields of natural language processing and information extraction, catering to a wide variety of linguistic and contextual product scenarios. The dataset's detailed annotations, encompassing item names, attributes, and categorization into 44 distinct product classes, provide a foundation for fostering breakthroughs in data processing and analytical applications.

Furthermore, we explored the capabilities of the fine-tuned LLaMA, harnessing the power of the LLaMA V1 and V2 models to achieve remarkable performance in the key information extraction and item classification tasks. These models, even in one-shot and few-shot learning scenarios, demonstrated superior results across numerous categories, illustrating their potential as tools for precise and efficient data extraction. 

Looking ahead, we aim to enhance AMuRD by incorporating additional features to extract finer details from receipts, thereby expanding its scope and utility. We also envision providing access to actual receipt images to further support diverse research avenues, such as the interpretation of textual information within images. Through these continuous improvements, we aspire for AMuRD to serve as a cornerstone resource, facilitating ongoing innovation and aiding researchers and practitioners in the evolving landscape of information extraction.
\newpage
\bibliographystyle{ACM-Reference-Format}
\bibliography{sample-base}

\end{document}